\documentclass{article}

\usepackage{arxiv}

\usepackage[utf8]{inputenc} 
\usepackage[T1]{fontenc}    
\usepackage{hyperref}       
\usepackage{url}            
\usepackage{booktabs}       
\usepackage{amsfonts,amsmath,amsthm}       
\usepackage{nicefrac}       
\usepackage{microtype}      
\usepackage{lipsum}
\usepackage{graphicx}

\newtheorem{proposition}{Proposition}

\title{New Spiking Architecture for Multi-Modal Decision-Making in Autonomous Vehicles}

\author{
 Aref Ghoreishee, Abhishek Mishra, Lifeng Zhou, John Walsh, and Nagarajan Kandasamy \\
  Electrical and Computer Engineering Department\\
  Drexel University, Philadelphia, PA 19104\\
}

\begin{document}
\maketitle

\begin{abstract}
This work proposes an end-to-end multi-modal reinforcement learning framework for high-level decision-making in autonomous vehicles. The framework integrates heterogeneous sensory input, including camera images, LiDAR point clouds, and vehicle heading information, through a cross-attention transformer-based perception module. Although transformers have become the backbone of modern multi-modal architectures, their high computational cost limits their deployment in resource-constrained edge environments. To overcome this challenge, we propose a spiking temporal-aware transformer-like architecture that uses ternary spiking neurons for computationally efficient multi-modal fusion. Comprehensive evaluations across multiple tasks in the Highway Environment demonstrate the effectiveness and efficiency of the proposed approach for real-time autonomous decision-making.
\end{abstract}




         
\newcommand{\BibTeX}{\rm B\kern-.05em{\sc i\kern-.025em b}\kern-.08em\TeX}
 

\section{Introduction}
Perception is the cornerstone of modern autonomous agents, transforming raw sensory data into representations suitable for high-level decision making. In real-world scenarios, agents must process heterogeneous data streams, such as camera images, LiDAR point clouds, and IMU signals, and fuse them into a coherent and task-relevant understanding of the environment~\cite{huang2202multi, li2020deep, sun2020scalability}. Recent work has explored mechanisms for cross-attention~\cite{nagrani2021attention, wei2020multi}, contrastive learning~\cite{zhou2024vision}, and representation alignment~\cite{jin2025multimodal} to maximize the informativeness and reliability of such fused representations. Among these techniques, cross-attention has emerged as the backbone of modern perception systems, allowing the effective integration of complementary modalities for complex tasks such as autonomous navigation, where the accuracy of decisions is critically dependent on perception quality~\cite{nagrani2021attention, wei2020multi, helvig2024caff, shen2024icafusion}.

Reinforcement learning (RL) allows agents to learn complex decision-making strategies in dynamic and uncertain environments through direct interaction with the world. \emph{Deep reinforcement learning} (\emph{DRL}), which integrates deep neural networks to process high-dimensional sensory observations with RL approaches, has achieved impressive results in autonomous driving and robotic navigation~\cite{zhu2017target, ha2018automated, hu2021sim, zhang2021tactical}. Since decision-making performance is highly dependent on perception quality, integrating attention-based fusion modules into end-to-end DRL frameworks has become increasingly essential~\cite{chen2024end}. Compared to traditional pipelines with separate, hand-engineered perception and control stages, \emph{end-to-end learning} jointly optimizes both components, enabling the development of robust, task-specific representations and policies directly from raw sensor inputs~\cite{chen2020end, lu2025hierarchical}. However, the high computational demands of DRL with attention-based fusion remain a major barrier to the deployment of such systems in autonomous agents with limited hardware resources~\cite{hu2024high, tang2021deep}. \emph{Spiking Neural Networks} (\emph{SNNs}) offer a promising alternative by providing an event-driven, computationally efficient paradigm inspired by biological neurons~\cite{cao2015spiking, wu2018spatio, lee2020enabling, lv2022spiking}. SNNs transmit information through discrete spikes, where sparse neuronal activity and the dominance of addition operations over costly multiplications lead to highly efficient computation~\cite{naveros2014spiking, teeter2018generalized}. These properties make SNNs attractive for resource-limited autonomous systems, but integrating attention mechanisms into spiking architectures introduces new challenges.

Attention mechanisms, such as the one proposed in the transformer architecture~\cite{vaswani2017attention}, calculate similarity scores using the dot products of the query-key pairs followed by a softmax normalization, which requires multiplications and exponentiations. These operations are incompatible with the multiplication-free, event-driven design principles of SNNs and incur significant computational costs~\cite{zhou2022spikformer}. To address this, recent work has proposed \emph{spiking self-attention} (\emph{SSA}) mechanisms, where Query, Key, and Value vectors are encoded as binary or a combination of binary and ternary spikes, enabling attention maps to be computed without softmax normalization~\cite{zhou2022spikformer, guo2025spiking}. This approach substantially reduces computational complexity and has shown promising results in image classification tasks.

Since the self-attention mechanism can be adapted for cross-attention by assigning the Query, Key, and Value to different modalities, a similar approach can be applied in the spiking domain. \emph{However, existing SSA models disregard temporal dependencies between spike-encoded signals, even though spike encoding naturally introduces a temporal dimension into the system.} To overcome this limitation, we propose a temporal-aware spiking cross-attention mechanism based on ternary spiking neurons that explicitly captures spatial dependencies while maintaining temporal dependencies, computational efficiency, and enhancing representational capacity. Building on this module, we design an end-to-end Q-learning architecture for autonomous driving in which the perception stage fuses multiple modalities using event-driven multiplication-free attention. Combined with SNNs, the proposed architectures achieve competitive decision-making performance with significantly reduced computational cost. To the best of our knowledge, this is the first work to integrate spiking cross-attention-based fusion into end-to-end reinforcement learning for autonomous driving.

The main contributions of this work are summarized as follows:
\begin{itemize}
    \item \emph{End-to-end multi-modal Q-learning network architecture (MM-DQN)}, a cross-attention-based fusion framework that integrates bird’s-eye view (BEV) images, and LiDAR and IMU data for autonomous driving decision-making.
    
    \item \emph{Temporal-aware ternary spiking attention (TTSA)}, a new attention mechanism that explicitly maintains temporal and sign dependencies when calculating the attention map, thus addressing a key limitation in prior SSA models.
    
    \item \emph{Comprehensive experimental validations} demonstrate that multi-modal reinforcement learning achieves a better performance with fewer successive observations compared to single-modal RL. Moreover, while earlier spiking attention mechanisms exhibit a performance gap compared to non-spiking models, our proposed temporal-aware spiking attention substantially reduces this gap. The corresponding computational cost is also quantified. 
\end{itemize}

The remainder of the paper is organized as follows. Section~\ref{sec:related_work} discusses related work, and Section~\ref{sec:preliminaries} introduces the basic concepts underlying Deep-Q and spiking neural networks. Section~\ref{sec:MM_DQN} develops the non-spiking MM-DQN architecture, while Section~\ref{sec:TTSA} develops the corresponding spiking version using the TTSA mechanism. Section~\ref{sec:experiments} evaluates the performance of the various architectures. We conclude the paper in Section~\ref{sec:conclusions}.   

\section{Related Work}\label{sec:related_work}
We discuss related work in the area of multimodal RL and the use of SNNs to develop the attention mechanisms required to fuse multiple modalities of information. 

\subsection{MultiSensor and Multimodal RL}
Chen et al. propose an RL framework for multisensor agents (vision, touch, proprioception) to maximize the shared information between modalities~\cite{chen2021multi}. Noting that naive multisensor RL often over-relies on a single modality or learns inconsistent representations when inputs are noisy or missing, the authors introduce a latent state-space model trained with a mutual information objective that explicitly aligns embeddings across modalities. In robotics, Becker et al. tackle the challenge of learning effective state representations from heterogeneous inputs (high-dimensional visual data and low-dimensional signals) by applying distinct self-supervised losses to each modality, respecting their differing information content~\cite{becker2023combining}. Jangir et al. develop a transformer-based RL framework that fuses a static third-person camera with a robot-mounted egocentric camera using cross-view attention, allowing features from one view to query and fuse with the other before feeding into a soft-actor-critical policy~\cite{jangir2022look}. Su et al. propose an algorithm for autonomous driving, building a spatio-temporal graph of nearby vehicles and combining it with camera images through cross-attention~\cite{su2024multimodal}. Finally, Lu et al. develop a deep RL system for adaptive scheduling in global logistics that integrates heterogeneous IoT sensor data (RFID, GPS, and environmental sensors) with external context (such as traffic, weather) through cross-attention, aligning, and weighting modalities before passing them to an actor-critic policy for decision optimization under dynamic conditions~\cite{lu2025multimodal}.

Despite these advances, performing multimodal fusion for perception remains computationally expensive, posing challenges for autonomous systems with limited processing resources. SSNs offer a promising path toward more efficient multimodal RL. 

\vspace{-6pt}

\subsection{Spiking Neural Networks}
SNNs emulate biological neurons by transmitting information via discrete spikes rather than continuous signals. Foundational models such as Hodgkin–Huxley~\cite{hodgkin1952currents} and Leaky Integrate-and-Fire~\cite{maass1997networks} laid the groundwork for simulating neuronal dynamics, while Spike-Timing-Dependent Plasticity~\cite{caporale2008spike} enabled the learning of temporal patterns. Many ANN architectures can be adapted to SNNs by substituting activation functions with spiking neurons, and recent research explores two main training paradigms: ANN-to-SNN conversion, which maps pre-trained ANNs to SNNs but often suffers from high latency due to long simulation windows~\cite{bu2022optimized, hao2023reducing, ho2021tcl, sengupta2019going}, and surrogate gradient learning, which trains SNNs directly via backpropagation, achieving high accuracy with fewer time steps and lower latency~\cite{fang2021deep, guo2024enof, guo2023rmp, guo2023joint, li2021differentiable, wu2018spatio, zhang2021rectified, liu2022human}. 

However, attention mechanisms, which rely on dot products and softmax operations involving multiplications, divisions, and exponentials, remain incompatible with the computing model underlying SNNs, requiring the development of novel attention architectures specifically designed for spiking computation.

\vspace{-6pt}

\subsection{Spike-Based Attention Mechanisms}
Zhou et al. introduce a spiking self-attention mechanism to align self-attention with the SNN's computing model by encoding Queries, Keys, and Values as spikes (0/1) and directly computing attention through dot products, removing softmax~\cite{zhou2022spikformer}. In a similar vein, other authors have also proposed spiking architectures that replace costly operations such as multiplications and softmax with simpler addition operations to improve computational efficiency in SNNs~\cite{yao2023spike,guo2025spiking}. Guo et al. introduced a spiking approach to self-attention using binary spikes for Queries, ReLU for Keys, and ternary spikes for Values to eliminate softmax while improving representational capacity~\cite{guo2025spiking}.

Previous work mostly ignores temporal dependencies introduced by spike encoding when calculating the attention map, while primarily focusing on self-attention. To address this limitation, we tackle the spiking cross-attention problem, critical for multi-sensor data fusion, and propose a novel spiking attention mechanism that captures temporal dependencies between encoded spikes. 


\section{Preliminaries}\label{sec:preliminaries}
This section introduces the basic concepts underlying Deep-Q networks and spiking neural networks.
\vspace{-6pt}

\subsection{Deep-Q Network}
The goal of \textit{RL} is to learn a policy $\pi$ that maximizes the expected cumulative return $\mathbb{E}[R(\tau)]$ in a Markov decision process, defined by the tuple $(\mathcal{S}, \mathcal{A}, P, R)$, where $s \in \mathcal{S}$ are states, $a \in \mathcal{A}$ are actions, $P(s'|s,a)$ denotes the transition dynamics and $r = R(s,a)$ is the reward function. In \textit{Deep Q-Learning}, the policy is implicitly defined through the action-value function $Q_\theta(s,a)$, parameterized by $\theta$, which approximates the expected return starting from state $s$, taking action $a$, and thereafter following the optimal policy  
\begin{equation}
Q^*(s,a) = \max_\pi \mathbb{E} \left[ \sum_{t=0}^{\infty} \gamma^t r_t \,\middle|\, s_0 = s, a_0 = a, \pi \right].
\end{equation}
The optimal Q-function satisfies the Bellman optimality equation  
\begin{equation}
Q^*(s,a) = \mathbb{E}_{s'} \left[ r + \gamma \max_{a'} Q^*(s',a') \right],
\end{equation}
which is approximated in a Deep Q-network (DQN) by minimizing the temporal difference error as  
\begin{equation}
\mathcal{L}_{\theta} = \mathbb{E}_{(s,a,r,s') \sim D} \Big[ \big( r + \gamma \max_{a'} Q_{\theta^-}(s',a') - Q_\theta(s,a) \big)^2 \Big],
\end{equation}
where $D$ is a replay buffer that stores transitions $(s,a,r,s')$ and $Q_{\theta^-}$ is a periodically updated target network for stable learning~\cite{li2023reinforcement}. The learned policy is then given by $\pi(s) = \arg \max_a Q_\theta(s,a)$.

\vspace{-6pt}

\subsection{Spike Encoding}\label{sec:spike_encoding}
Unlike conventional deep neural networks, SNNs process sparse spike trains rather than continuous signals. To enable this, each real-valued input is converted into a spike train using \textit{spike encoding} in which every data point is expanded into a sequence of spikes $(0, 1,$ or sometimes $-1)$ over a simulation time window, which sets the maximum number of possible spikes. This naturally introduces a temporal dimension into the representation as each spike occurs at a specific time step. The interval between possible spike events defines the time step, and lengthening the simulation window directly increases network latency, since the output can only be computed after the full spike sequence is processed.

A widely used encoding strategy is \emph{rate encoding} that represents the strength of the signal by the spike rate (widely observed in sensory systems such as the visual, motor, and auditory cortices). The mapping of normalized signals to firing rates can be direct or probabilistic~\cite{bian2024evaluation}. For example, Poisson encoding generates spikes stochastically with probabilities proportional to input intensity, while Bernoulli encoding uses independent Bernoulli trials at each time step with spike probabilities determined by the signal value. The resulting spike trains then propagate through the SNN and the output spikes are aggregated to produce the final result.

\vspace{-6pt}

\subsection{Spiking Neural Networks}
Spiking neurons are the fundamental computational units of SNNs, integrating spatial and temporal information into the membrane potential, which is then transformed into binary spikes $(0,1)$ or ternary spikes $(0,1,-1)$. These spikes propagate through successive layers of the SNN until the output layer is reached~\cite{ghoreishee2025improving}. Neuron dynamics is modeled using the \textit{Leaky Integrate-and-Fire (LIF)} model~\cite{maass1997networks}, where the evolution of the membrane potential and the spike generation mechanism are described as:
\begin{align}
m(t) &= v(t-1) + x(t), \nonumber \\
s(t) &= F_{\text{spk}}\big(m(t)\big), \nonumber \\
v(t) &= \beta\, m(t)\big(1 - s(t)\big) + v_{\text{reset}}\, s(t).
\label{Eq5-7}
\end{align}
Here, $x(t)$ denotes the spatial input current at time $t$ from the previous layer, $m(t)$ is the membrane potential combining the spatial input $x(t)$ with the temporal input $v(t-1)$ from the previous time step, $F_{\text{spk}}(\cdot)$ is the spike-generation function, $s(t)$ is the spike output at time $t$, $v_{\text{reset}}$ is the reset potential, and $\beta$ is the factor controlling the decay in membrane potential. The spike-generating function for the binary spiking neuron is defined as
\begin{equation}
F_{\text{spk}}\big(m(t)\big) = H\big(m(t) - V^{\text{th}}\big),
\label{Eq8}
\end{equation}
and for the ternary spiking neuron as
\begin{equation}
F_{\text{spk}}\big(m(t)\big) = H\big(m(t)-V^{\text{th}_p}\big) - H\big(m(t)-V^{\text{th}_n}\big),
\label{eq9}
\end{equation}
where $H(\cdot)$ is the Heaviside step function. In the binary case, when the membrane potential $m(t)$ exceeds the threshold $V^{\text{th}}$, the neuron fires ($s(t)=1$) and resets to $v_{\text{reset}}$; otherwise, $m(t)$ decays at a rate determined by $\beta$ to form the temporal input for the next time step. For ternary neurons, a positive spike is generated as in the binary case, while $m(t)$ below the negative threshold $V^{\text{th}_n}$ produces a negative spike ($s(t)=-1$) with the same reset behavior; otherwise, $m(t)$ decays at the same rate to form the next temporal input.

For notational simplicity, we denote the binary and ternary spiking neuron layers as $\text{BSN}(\cdot)$ and $\text{TSN}(\cdot)$, respectively, where the input is the output of the previous layer and the output is the spike tensor $s$. In the output layer, the spike trains across the entire simulation window are summed to generate the final output values.

\section{Multi-Modal DQN Architecture}\label{sec:MM_DQN}
We describe our first contribution: an end-to-end multi-modal DQN (MM-DQN) that uses a cross-attention module to fuse BEV images, and LiDAR and IMU data for high-level decision-making in autonomous driving. 

\begin{figure}[!t] 
  \centering
  \includegraphics[width=0.98\linewidth]{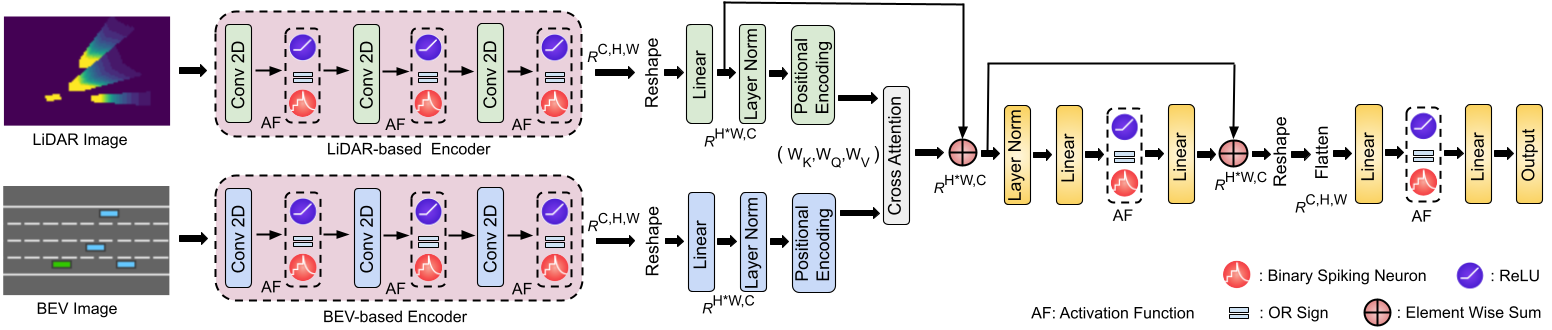}
  \caption{The MM-DQN architecture that uses a cross-attention module to fuse BEV images and LiDAR data for RL-based decision making. The figure also shows the choice between ReLU and spiking activations; in the traditional ANN, ReLU activations are used throughout, whereas all ReLU activations are replaced by binary spiking neurons in case of the SNN.}
  \label{fig1}
\end{figure}

\begin{figure}[!t] 
  \centering
  \includegraphics[width=0.60\linewidth]{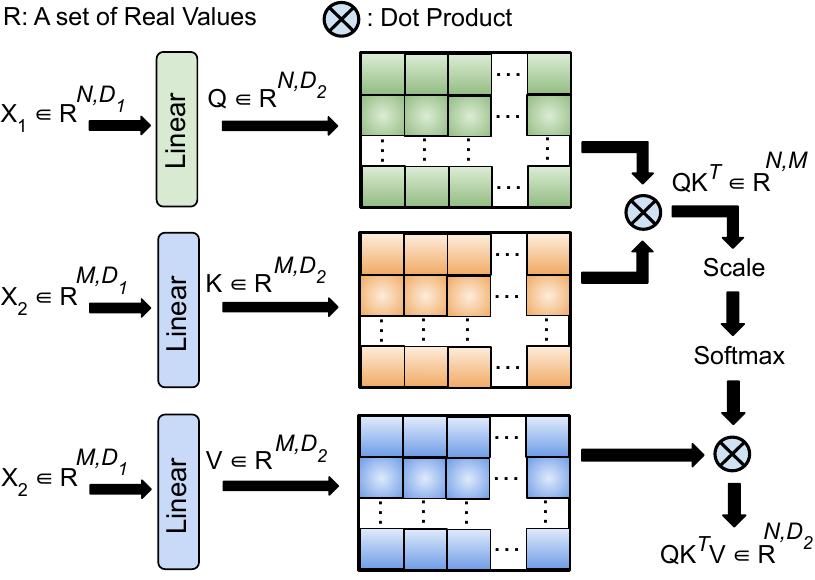}
  \caption{Transformer-like cross-attention mechanism.}
  \label{fig2}
  \vspace{-18pt}
\end{figure}

\subsection{Overall Architecture}
Figure~\ref{fig2} shows the MM-DQN architecture comprising four main components: a convolution-based feature extractor, feature embedding, cross-fusion, and a fully connected decision head.

The two inputs are denoted as $X_1, X_2 \in \mathbb{R}^{C \times H \times W}$, where $C$ is the number of channels, and $H, W$ represents the height and width of the images. Although the two modalities may differ in resolution, for notational simplicity, we use the same symbol for their dimensions. The channel dimension can represent standard RGB channels for visual images or multiple consecutive frames stacked along the channel axis~\cite{mnih2015human}. Sensory data are preprocessed into two image-based input modalities: (1) BEV representation of the scene, obtained by transforming images from calibrated vehicle-mounted cameras into a unified top-down view using geometric projection or learned depth estimation methods~\cite{li2024bevformer, philion2020lift}; and (2) LiDAR point cloud data, converted into a pseudo-image representation by projecting 3D points onto 2D grids for top-down spatial mapping~\cite{lang2019pointpillars}. These LiDAR projections can encode attributes such as height, intensity, relative velocity, and density in pixel values. 

Since the input images are high-resolution, directly applying pixel-wise cross-attention between them is computationally expensive. To mitigate this, we use two convolution-based feature extraction modules, each consisting of multiple convolutional layers that progressively down-sample the spatial resolution while expanding the channel dimension, thereby reducing the height and width of the feature maps and simultaneously enriching their semantic representation. The transformed feature representations are obtained as $X_1' = \mathrm{CONV}(X_1)$ and $ X_2' = \mathrm{CONV}(X_2)$, where $X_1', X_2' \in \mathbb{R}^{c \times h \times w}, \quad h < H, \; w < W, \; c > C$. The \(\mathrm{CONV}(\cdot)\) block consists of stacked convolutional layers (with ReLU activations in the ANN case, and binary spiking neurons for the SNN variant). Each pixel in $X_1'$ and $X_2'$ is then treated as a $c$-dimensional feature vector (analogous to a ``word'' in NLP). The feature embedding module reshapes the feature maps into $h \times w$ vectors, each of dimension $c$, and applies a linear projection to map each vector into the embedding space $E_{X_1} = \mathrm{EMB}(X_1')$ and $E_{X_2} = \mathrm{EMB}(X_2')$, where $E_{X_1}, E_{X_2} \in \mathbb{R}^{h \times w \times c}$ and $\mathrm{EMB}(\cdot)$ involves a reshaping operation followed by a learnable linear layer.

The embedded features \(E_{X_1}, E_{X_2}\) are then passed through the cross-fusion module $\mathrm{CFL}(E_{X_1}, E_{X_2})$, which, inspired by Vaswani et al.~\cite{vaswani2017attention}, uses a Transformer-like architecture with a cross-attention layer (described in the next section). The CFL contains a multi-head cross-attention sublayer and a feed-forward sublayer. Both sublayers use residual connections along with layer normalization, with each output calculated as $\mathrm{LayerNorm}(x + \mathrm{Sublayer}(x))$, where $\mathrm{Sublayer}(x)$ is the function implemented by the sublayer and $x$ is its input. The cross-fusion output is given by $F_{x_1,x_2} = \mathrm{CFL}(E_{X_1}, E_{X_2}), \quad F_{x_1,x_2} \in \mathbb{R}^{h \times w \times c}$. 

Finally, $F_{x_1,x_2}$ is reshaped into $\mathbb{R}^{c \times h \times w}$, flattened and processed by the fully connected decision head comprising fully connected linear layers to generate the action-value function for each possible action $Q((X_1,X_2), a) = \mathrm{FC}(\mathrm{flatten}(F_{x_1,x_2}))$. Here, $(X_1,X_2)$ jointly form the state observation for the RL agent.

\subsection{Cross-Attention Mechanism}
The attention mechanism used in MM-DQN and shown in Fig.~\ref{fig2}, is a variant commonly used in models such as Transformers, designed to let one set of inputs (queries) attend to another set of inputs (keys and values). The cross-attention operation is expressed as
\begin{equation}
\text{CA}(Q,K,V) = \text{softmax}\left(\frac{QK^T}{\sqrt{d_k}}\right)V,
\label{eq10}
\end{equation}
where $Q$, $K$, and $V$ denote the query, key, and value matrices, respectively, and $d_k$ is the dimension of the key vectors. The $\text{softmax}$ function normalizes the attention scores, producing a weighted sum of the values based on the similarity between queries and keys.

The inputs to the cross-attention mechanism consist of a sequence of embeddings, $X_1 \in \mathbb{R}^{N \times d_1}$ for the query and $X_2 \in \mathbb{R}^{N \times d_1}$ for the key and value. These are linearly projected onto $Q$, $K$, and $V$ using learned-weight matrices $W_Q$, $W_K$, and $W_V$ corresponding to the query, key, and value projections, respectively, as
\begin{equation}
Q = X_1W_Q \quad K = X_2W_K \quad V = X_2W_V.
\end{equation}

\section{Multi-Modal Spiking Architecture} \label{sec:TTSA}
The previously developed MM-DQN is a traditional architecture incompatible with the spike-based computing model. This section describes our second contribution: the development of a spiking variant using a temporal-aware spiking attention mechanism that explicitly maintains temporal dependencies when calculating the cross-attention map for multimodal fusion.
\vspace{-6pt}

\subsection{Overall Architecture}
To obtain the SNN variant of MM-DQN, the input is first encoded as spike trains using rate-based encoding (see Section~\ref{sec:spike_encoding}). Converting the CONV-based feature extractor, feature embedding, and fully connected decision head layers is straightforward: ReLU activation functions can be directly replaced with spiking neuron models (e.g., LIF neurons). However, this simple substitution is not directly applicable to the calculations needed to generate the cross-attention map since these involve floating-point multiplications and softmax calculations. Therefore, recent work has proposed various specialized mechanisms to achieve a spike-based attention calculation~\cite{zhou2022spikformer, yao2023spike, guo2025spiking}. 

\begin{figure}[!t] 
  \centering
\includegraphics[width=0.55\linewidth]{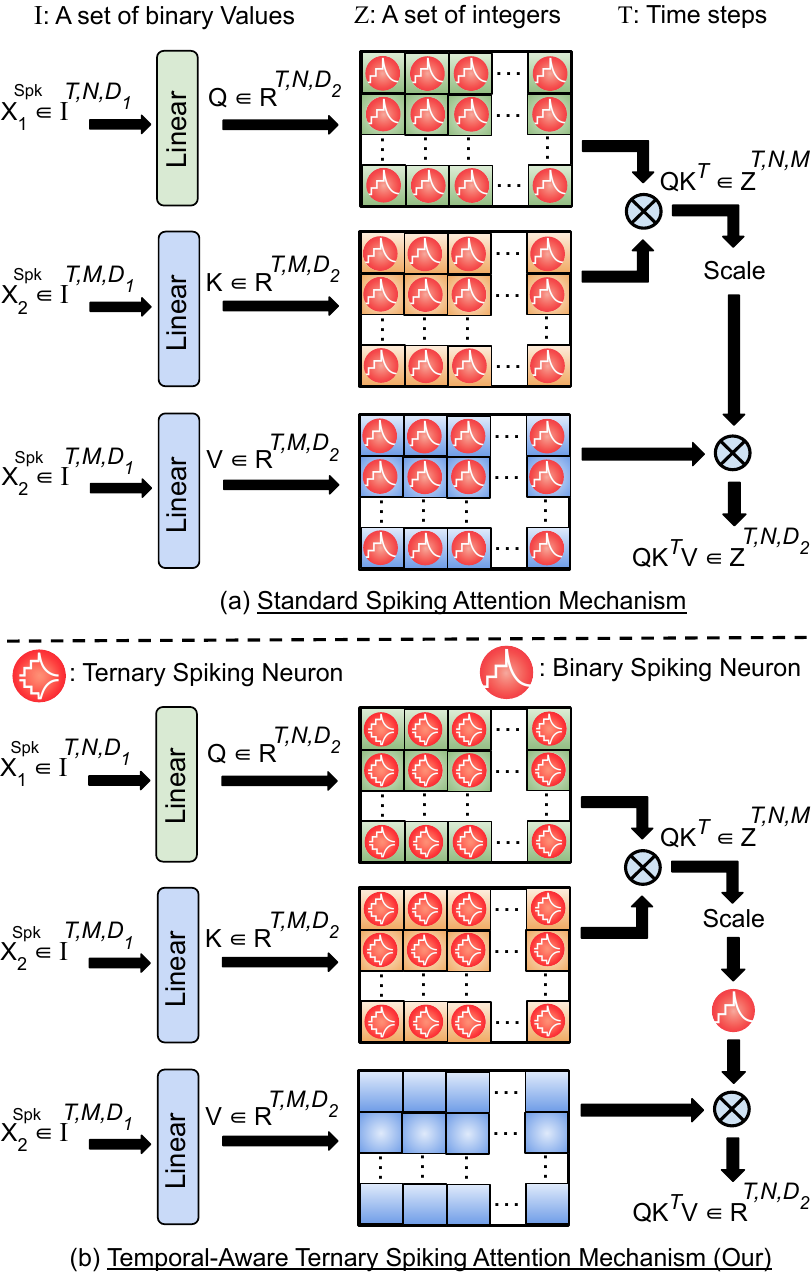}
  \caption{The standard spiking attention mechanism compared with our proposed temporal-aware approach.}
  \label{fig: Spk_attention}
  \vspace{-12pt}
\end{figure}

Zhou et al. proposed a spiking mechanism to achieve self-attention that can be extended to cross-attention by using different inputs for queries, keys, and values~\cite{zhou2022spikformer}. This approach is shown in Fig.~\ref{fig: Spk_attention}(a), where $Q$, $K$, and $V$ are first obtained via learnable weight matrices and then converted into spiking sequences through binary spiking neurons as 
\begin{equation}
Q = BSN_Q \bigl( BN(X_1W_Q) \bigr) \quad K = BSN_K \bigl( BN(X_2W_K) \bigr) \quad V = BSN_V \bigl( BN(X_2W_V) \bigr)
\end{equation}
where $BN(\cdot)$ denotes batch normalization and $BSN(\cdot)$ denotes the binary spiking neuron layer. The attention matrix is then calculated using queries and keys in the form of binary spikes (containing only $0$s and $1$s) and, therefore, replacing matrix multiplications with addition operations. A scaling factor $\omega$ controls the magnitude of the result, leading to the formulation
\begin{equation}
SSA(Q,K,V) = BSN\bigl(QK^T V\omega\bigr),
\label{eq15}
\end{equation}
where $\omega$ regulates the values of the matrix product, while all other calculations, including obtaining the attention map, are performed using addition, in line with the spike-driven computing paradigm. 

We refer to the architecture described above as \textit{Standard Spiking Attention (SSA)} to differentiate it from our proposed approach. Although SSA uses spike-based computing to calculate the attention map (by converting queries, keys, and values into binary spikes before performing matrix multiplication), this typically leads to significant information loss that degrades result quality. In the following sections, we analyze this issue in detail and propose a method to mitigate this loss of information.  
\vspace{-6pt}

\subsection{Analysis of Information Loss}\label{sec:issue}
We present a detailed mathematical analysis of the information loss inherent to the SSA mechanism. Encoding queries, keys, and values as 0s and 1s using binary spiking neurons reduces representational capacity, inevitably leading to information loss. To formalize this claim, we employ an \emph{entropy-based analysis}. The maximum representational capacity of a set $X$ is its maximum entropy
\begin{equation}
    C(X) = \max H(X) = \max \Bigl(- \sum_{x \in X} p_X(x) \log p_X(x)\Bigr),
\end{equation}
where $p_X(x)$ denotes the probability distribution over $X$.  

\begin{proposition}
The representational capacity $C(X)$ achieves its maximum when $X$ follows a uniform distribution, i.e., $p_X(x) = 1/N$, where  $N$ is the number of samples. Under this condition,
\begin{equation}
    C(X) = \log N.
\label{eq17}
\end{equation}
\label{propos1}
\end{proposition}

\vspace{-18pt}

\noindent Proposition~\ref{propos1} allows us to compare the representational capacity of SSA \eqref{eq15}, which uses binary spike-encoded queries, keys, and values, to the traditional, non-spiking attention mechanism \eqref{eq10}. Let $ X \subset \mathbb{R}^{c \times h \times w} $ denote the floating-point values and $S\subset \{0,1\}^{T \times c \times h \times w}$ the binary spike-encoded values with $ S = BSN(X) $.  For 32-bit precision floating-point values we obtain 
\begin{equation}
    C(X) = \log \bigl(2^{32 \times c \times h \times w}\bigr) = 32 \times c \times h \times w.
\end{equation}
whereas in the case of binary spikes using 1-bit representation, 
\begin{equation}
    C(S) = \log \bigl(2^{T \times c \times h \times w}\bigr) = T \times c \times h \times w,
\label{eq19}
\end{equation}
where $T$ is the number of time steps. However, in SNNs, only the current time step  $s(t) \in \{0,1\}^{c \times h \times w}$ is available at each computation stage. Thus, the effective representation capacity per step is further reduced to
\begin{equation}
    C(S(t)) = c \times h \times w,
\end{equation}
leading to a significant information loss compared to the case of using floating-point values.

\emph{It is important to note that a significant part of this information loss arises from the sign bit.} Recall from \eqref{Eq5-7} and \eqref{Eq8}, that all negative floating-point values are encoded as zeros, regardless of their magnitude. The magnitude of positive values is further encoded temporally (number of spikes in the simulation time window), but since SNNs process only a single time step at once, this temporal information is partially discarded. We formalize this observation through the following proposition.

\begin{proposition}
Binary spike encoding destroys negative--negative alignment and temporal information. Consider two values $q, k \in \mathbb{R}^d$. For each dimendion $i$, a binary spiking neuron generates a spike train
\begin{equation}
    s_i(t; x) \in \{0,1\}, \qquad t = 1,\dots,T,
\end{equation}
where $x$ represents either $q$ or $k$. Applying the nonnegativity property \eqref{Eq8}, we obtain
\begin{equation}
    \forall x \le 0: \quad s_i(t; x) = 0 \quad \forall t,
\end{equation}
i.e., nonpositive inputs produce no spikes. Define the attention map as 
\begin{equation}
    M(q,k) = \sum_{i=1}^d \sum_{t=1}^T s_i(t; q_i) \, s_i(t; k_i).
\end{equation}
Then:
\begin{enumerate}
    \item The set \( S_{--} := \{ i : q_i < 0,\, k_i < 0 \} \) contributes zero to \( M(q,k) \).
    \item The real-valued dot product decomposes as
    \begin{equation}
        q^\top k = \sum_{i \in S_{++}} |q_i||k_i| + \sum_{i \in S_{--}} |q_i||k_i| - \sum_{i \in S_{+-} \cup S_{-+}} |q_i||k_i|,
    \end{equation}
    where \( S_{++} := \{ i : q_i > 0, k_i > 0 \} \), etc. Hence, \( S_{--} \) contributes positively to \( q^\top k \).
    \item Therefore, there exist $q, k$ such that  $q^\top k > 0$ but $M(q,k) = 0$.
\end{enumerate}
\label{propos2}
\end{proposition}
As a consequence of Proposition~\ref{propos2}, SSA suffers from two main types of information loss:
\begin{enumerate}
    \item \textbf{Negative--negative alignment loss}: contributions from \( S_{--} \) are discarded.
    \item \textbf{Temporal loss}: only the current time step is considered, ignoring the magnitude information encoded over time.
\end{enumerate}

Together, these factors explain the significant representational loss in SSA compared to the more real-valued attention mechanism used in MM-DQN. 

\vspace{-6pt}

\subsection{Temporal-Aware Ternary Spiking Attention}
Our method reduces the information loss described above using a \textit{Temporal-Aware Ternary Spiking Attention} (\emph{TTSA}) mechanism, shown in Fig.~\ref{fig: Spk_attention}(b).

To reduce the negative--negative alignment loss when calculating the attention map using spike-based computing while preserving the \textit{multiplication-free property} TTSA uses the ternary spiking neuron, previously defined in \eqref{eq9}, to encode the query and key representations. Ternary spiking neurons encode negative inputs as $-1$ and positive inputs as $+1$, the number of spikes depending on the magnitude of the inputs. Consequently, ternary spikes use two bits to represent $\{-1, 0, 1\}$, allowing for increased information capacity compared to binary spikes. Using \eqref{eq17}, we obtain
\begin{equation}
C(S(t)) = 2 \times T \times c \times h \times w,
\label{eq:capacity_ternary}
\end{equation}
and the representational capacity of the ternary spikes exceeds that of the binary spikes obtained in \eqref{eq19}.  

Returning to Fig.~\ref{fig: Spk_attention}(b), the process for encoding values for the query and key in TTSA is as follows:
\begin{equation}
Q = TSN(BN(X_1 W_Q)) \qquad K = TSN(BN(X_2 W_K)).
\label{eq:qk_encoding}
\end{equation}
After obtaining the encoded query $Q$ and key $K$, the attention map is generated using the dot product $QK^\top$. Unlike the SSA mechanism that produces only nonnegative values in the attention map~\cite{zhou2022spikformer}, the proposed ternary spike-based encoding allows the attention map to include \textit{negative values}.  

Inspired by the softmax operation used in non-spiking networks such as MM-DQN to calculate the attention map, where negative elements in the attention map receive smaller weights (see \eqref{eq10}), we propose passing the attention map $QK^\top$ through a binary spiking neuron layer. This step serves two key purposes:
\begin{enumerate}
    \item \textbf{Suppressing negative correlations:} The binary encoding naturally assigns a weight of zero to negative elements in the attention map, similar to softmax reducing their contribution in the non-spiking attention mechanism.
    
    \item \textbf{Preserving temporal information:} From \eqref{Eq5-7} it follows that a fraction of the attention map at each time step is captured within the temporal state $v(t)$ of the neuron and propagated to subsequent steps. This allows the attention mechanism to capture temporal dependencies across spike sequences and therefore reduce some of the temporal loss discussed in the previous section.
\end{enumerate}

Since the attention map is already in spike form, unlike in SSA~\cite{zhou2022spikformer}, we do not apply any additional spike encoding to the $V$ matrix, and the binary-encoded attention map $BSN\bigl(QK^T\bigr)$ is used to weigh the value matrix $V$ as 
\begin{equation}
TTSA(Q,K,V) = BSN(QK^\top) \, V.
\label{eq:ttsa}
\end{equation}
Note that the entire calculation of the attention map does not require multiplications. 

\section{Experiments}\label{sec:experiments}
Our objectives are threefold: (1) to evaluate the performance of the proposed MM-DQN against its uni-modal counterparts; (2) to examine the performance gap between the spiking and conventional versions of MM-DQN; and (3) to assess the effectiveness of the proposed TTSA mechanism compared to the standard SSA in multi-modal fusion.

Experiments are conducted using Highway-Env involving \textit{highway} and \textit{roundabout} scenarios~\cite{leurent2018environment}. Highway-Env offers multiple sensory observations for the ego vehicle: grayscale BEV images, LiDAR, and kinematic data. BEV images represent sequential top-down views of the environment. The LiDAR observation divides the $360^\circ$ space around the vehicle into angular sectors and returns two values per sector: the distance to the nearest obstacle and its relative velocity along the beam direction. The result is an $N \times 2$ array that encodes both spatial and dynamic context. Kinematic data, equivalent to the IMU output, provide the velocity and heading of the ego vehicle. We convert LiDAR, velocity, and heading information into an image-like representation, where each pixel encodes the object position and its intensity corresponds to the relative velocity. More information on the simulation environment, data preprocessing steps, network architecture, and training and hyperparameter settings is provided in Appendices~\ref{app:A},~\ref{app:B},~and~\ref{app:c}.\footnote{The code is provided in the supplementary materials and is available via the link: \url{https://anonymous.4open.science/r/Spiking_MM_DQN-FFF3}.}

\vspace{-6pt}

\subsection{Uni-Modal versus Multi-Modal Performance}
Since the conventional MM-DQN employs CNN-based feature extractors, the well-known CNN-based uni-modal DQN~\cite{mnih2015human}, proposed in “Human-level control through deep reinforcement learning”, is used as the baseline for comparison. Inspired by ~\cite{mnih2015human}, the uni-modal model processes a stack of four successive BEV frames to extract temporal information such as object velocities through its CONV sublayer. In contrast, MM-DQN encodes key temporal cues, including velocity and ego-vehicle heading, directly into the LiDAR image representation. Consequently, it requires fewer successive frames for effective decision-making. MM-DQN therefore use a single BEV frame and a single LiDAR image representation as inputs. To analyze the effect of frame count, the uni-modal DQN is also trained using a single BEV frame.  

All models are trained in the \textit{highway} scenario for 100\,k steps using an $\epsilon$-greedy policy, with checkpoints saved every 5\,k steps. Training is repeated with five random seeds for robustness. Each saved policy is evaluated over 20 test scenarios and the average reward is reported. 

\begin{figure}[!t] 
  \centering
\includegraphics[width=0.45\linewidth]{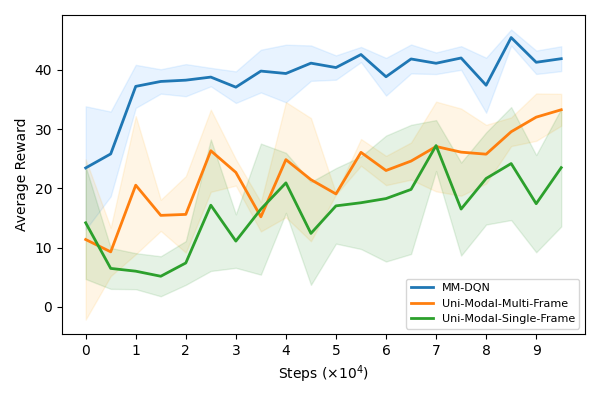} \vspace{-12pt}
  \caption{The average reward obtained by MM-DQN compared to uni-modal DQNs operating with a stack of four frames and a single frame.}
  \label{Result1}
  \vspace{-12pt}
\end{figure}

Figure~\ref{Result1} plots the average rewards obtained by MM-DQN and the uni-modal DQN. MM-DQN consistently outperforms the baseline. Each test scenario contains up to 50 steps, with a maximum achievable reward of 1 per step. The MM-DQN achieves an average reward of about 45, demonstrating its superior performance. The reduced performance of the uni-modal DQN with a single input frame highlights the importance of temporal stacking, though excessive frame stacking increases latency and memory costs, limiting real-time applicability. 

\vspace{-6pt}

\subsection{Conventional versus Spiking Performance}
SNNs use surrogate gradients to overcome the non-differentiability of spike generation, introducing approximation errors and potential convergence instability~\cite{ghoreishee2025improving, eshraghian2021training}. Their discrete and temporally sparse representations also reduce the precision of the encoding, contributing to the performance gap with conventional networks. In spiking DQNs, the action values are represented through spike activity within a limited simulation window. Although longer windows improve value discrimination, they increase latency, making real-time decision-making challenging. Consequently, SNNs often struggle to distinguish actions with similar Q-values, hindering near-optimal performance. In our experiments, we select the simulation time window equal to 5. 

As discussed in Section~\ref{sec:issue}, some of this performance gap arises from information loss in the standard spiking attention (SSA) mechanism. The proposed temporal-aware ternary spiking attention (TTSA) mitigates this issue while preserving the energy efficiency of spike-based computation. We evaluate three models: conventional MM-DQN, spiking MM-DQN with SSA, and spiking MM-DQN with TTSA—on the \textit{highway} and \textit{roundabout} scenarios of Highway-Env, using five random seeds and evaluating checkpoints saved every 5\,k steps.

We investigate the performance of spike-based versions of MM-DQN for two different tasks in the Highway-Env.

\subsubsection*{Highway Scenario}
\begin{figure}[!t] 
  \centering
  \vspace{-18pt}
    \includegraphics[width=0.45\linewidth]{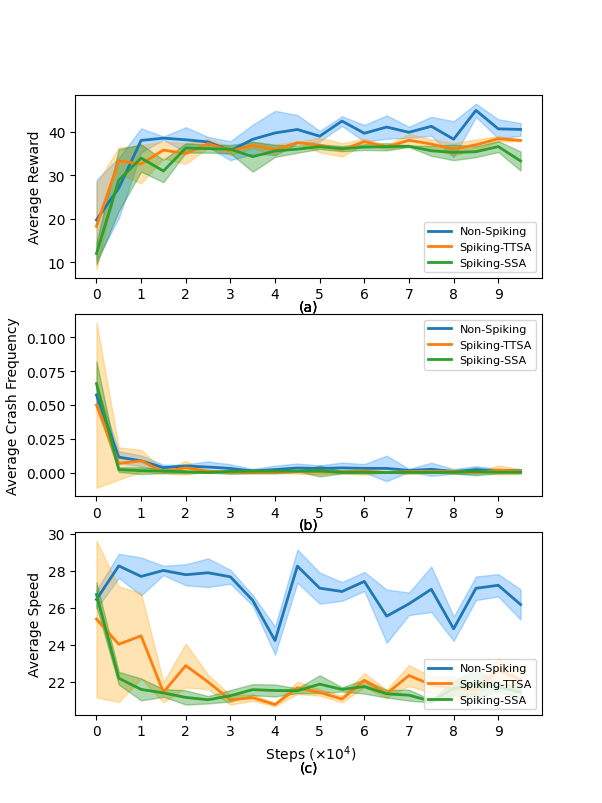}\vspace{-12pt}
  \caption{Average reward, crash frequency, and speed obtained by non-spiking, spiking with SSA, and spiking with TTSA architectures for the Highway scenario}
  \label{Result_3}
  \vspace{-12pt}
\end{figure}

The ego vehicle receives a positive reward for maintaining its speed within a predefined range, increasing with higher speeds, and a negative reward for collisions. The collision penalty and high-speed reward are set equally, encouraging faster driving. Figure~\ref{Result_3} presents the average reward, crash frequency, and speed across twenty test environments and five random seeds. Every 5{,}000 time steps, each model (five in total, one per seed) is evaluated over ten test runs to compute the average reward. 

The non-spiking agent achieves the highest average reward ($\approx 45$), while the spiking model with standard spiking attention (SSA) reaches $36.7$ ($\approx 18\%$ lower). The proposed TTSA-based spiking model improves this to $38.26$, narrowing the gap to $\approx 15\%$ and reducing the performance drop by about $3\%$. Crash frequencies are nearly identical across models. As shown in Fig.~\ref{Result_3}, the non-spiking agent maintains a higher average speed, suggesting a more aggressive policy, whereas the spiking variants exhibit slightly lower speeds and more conservative behavior.

\subsubsection*{Roundabout Scenario}
In this more complex scenario, the reward function combines multiple objectives, including collision avoidance, lane keeping, and smooth navigation. The default configuration assigns a stronger penalty for collisions ($-1$) than the reward for high speed ($0.2$). The environment is also more dynamic, as the heading and motion of nearby vehicles frequently change, making future positions harder to predict.

\begin{figure}[!t] 
  \centering
    \includegraphics[width=0.45\linewidth]{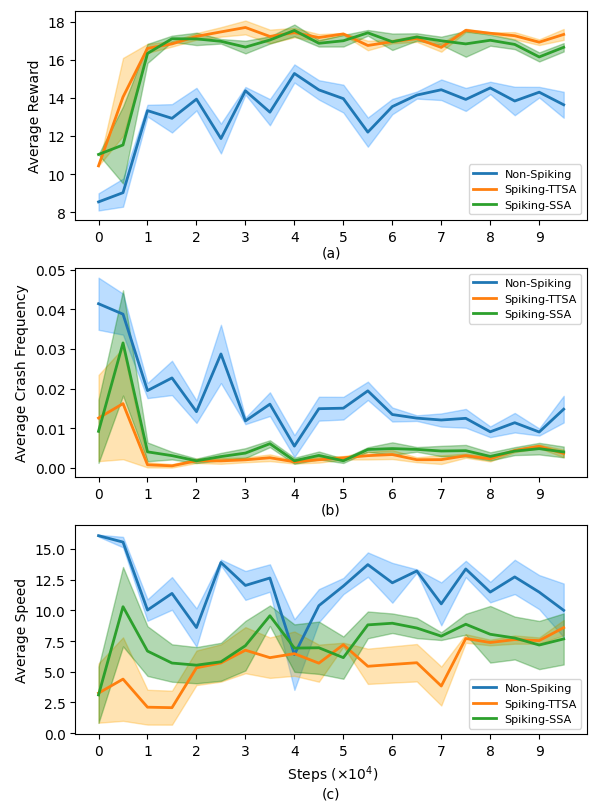}\vspace{-10pt}
  \caption{Average reward, crash frequency, and speed obtained by non-spiking, spiking with SSA, and spiking with TTSA architectures for the Roundabout scenario.}
  \vspace{-12pt}
  \label{Result4}
\end{figure}

We evaluate three metrics: the average reward, crash frequency (mean crashes per decision step), and average ego-vehicle speed. As shown in Fig.~\ref{Result4}, \emph{somewhat counterintuitively, the spiking variants achieve higher average rewards than the non-spiking model}. Figures~\ref{Result4}(b) and \ref{Result4}(c) clarify this behavior, spiking agents maintain lower speeds and substantially fewer crashes, reflecting a more cautious driving policy. 

Since the roundabout environment demands more precise decision-making, risky actions can yield higher rewards but also increase the likelihood of collisions. Similar to the highway scenario, SNNs tend to learn more conservative driving behaviors: spiking variants avoid risky yet potentially high-reward actions, resulting in safer but lower-reward policies. This is reflected in their lower speeds and reduced crash frequencies. The minimum and maximum average firing frequencies are $0.0017$--$0.031$ for the conventional MM-DQN, $0.005$--$0.038$ for the SSA-based MM-DQN, and $0.0005$--$0.016$ for the TTSA-based MM-DQN.  

A potential explanation is that sparse Q-value approximation in spiking variants mitigates Q-value overestimation, discouraging the selection of high-risk, high-reward actions. In contrast, the highway scenario, characterized by simpler and more predictable dynamics, where vehicle headings vary within a limited range, makes such risky actions more beneficial. This explains why less conservative (non-spiking) policies perform better in the highway scenario but worse in the roundabout setting, where the environment is more complex and stochastic.

Meanwhile, the spiking variant equipped with the proposed TTSA mechanism learns an even more cautious yet stable policy, achieving slightly higher average rewards than the SSA-based model and lower average crash frequency. This improvement suggests that TTSA enhances environmental understanding through better temporal attention and understanding of negative-negative alignment between data. 

\subsection{Computational Efficiency}
\begin{figure}[!t] 
  \centering
    \includegraphics[width=0.75\linewidth]{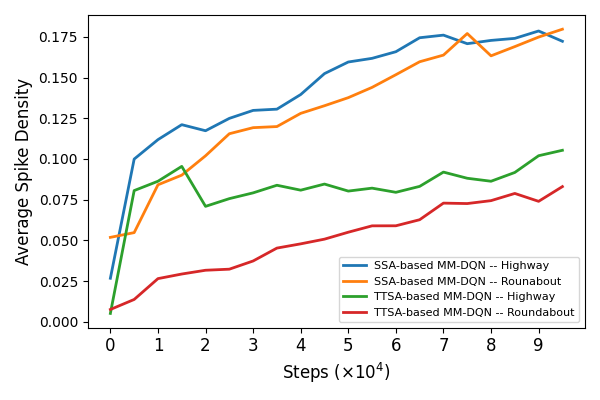}\vspace{-12pt}
  \caption{The average spike density observed by SSA-based MM-DQN and TTSA-based MM-DQN.}
   \label{Result7}
  \vspace{-15pt}
\end{figure}

Event-driven computation in SNNs means that neurons update their membrane potentials only when spikes occur, rather than at every time step as in conventional networks, the fundamental source of the superior energy efficiency of SNNs. We compare the average spike sparsity of the SSA-based MM-DQN and the proposed TTSA-based MM-DQN. Figure~\ref{Result7} presents the average spike density of the networks saved during training in the test environment. The TTSA-based MM-DQN exhibits \textbf{higher spike sparsity} (i.e., lower spike density) compared to the SSA-based MM-DQN, meaning fewer spike events and reduced computational activity. Despite its lower firing rate, the TTSA-based SNN achieves better performance. \emph{Since the only architectural difference between the two models lies in the attention mechanism, the improved sparsity can be attributed to the proposed TTSA module.} As discussed in Section~\ref{sec:issue}, this observation suggests that TTSA delivers a cleaner and more selective drive to downstream spiking layers, allowing neurons to fire only when it is truly informative to do so.

\section{Conclusions}\label{sec:conclusions}
We presented an end-to-end multi-modal deep Q-network (MM-DQN) with a cross-attention module for autonomous driving that efficiently fuses BEV and LiDAR modalities, outperforming uni-modal models while reducing memory and computational costs. A spiking variant was further developed for energy-efficient decision-making. To address the limitations of standard spiking attention, we proposed the Temporal-Aware Ternary Spiking Attention (TTSA) mechanism. Experiments on highway and roundabout scenarios show that TTSA-based SNNs maintain high decision stability and safety while achieving comparable or higher rewards than non-spiking baselines. TTSA also improves spike sparsity and reduces computational activity, narrowing the performance gap between spiking and conventional RL models. Future work will explore real-world deployment on neuromorphic and robotic platforms, and adaptive threshold mechanisms for enhanced temporal selectivity.

\clearpage

\bibliographystyle{unsrt} 
\bibliography{references}

\clearpage

\appendix

\section{Highway Env.}\label{app:A}
In all scenarios of the \texttt{Highway-Env} platform, the interaction between the reinforcement learning (RL) agent and the simulated driving environment is defined by the \textit{simulation frequency}, \textit{observation space}, and \textit{action space}. The simulation frequency determines how often the environment's physical dynamics are updated per second, while the policy frequency specifies how often the agent selects an action. Typically, the simulation operates at $f_s = 15~\text{Hz}$ and the agent makes decisions at $f_p = 1~\text{Hz}$, meaning several internal simulation steps occur between consecutive control decisions. This configuration ensures realistic vehicle motion.

\subsection{Observation Space}
The observation space provides the agent with the sensory and kinematic information required for perception and decision-making. \texttt{Highway-Env} supports multiple observation modalities that can be used independently or jointly within multi-modal reinforcement learning frameworks:

\paragraph{1) Kinematic Observation.}
The kinematic representation encodes the ego vehicle and nearby traffic as a matrix of dimension $V \times F$, where $V$ is the number of observed vehicles and $F$ is the number of features per vehicle. Each row contains continuous-valued attributes such as the longitudinal and lateral position, velocity, heading, and lane index:
\begin{equation}
o_t = [x_i,\, y_i,\, v_{x_i},\, v_{y_i},\, \text{heading}_i]_{i=1}^{V}.
\end{equation}
To ensure a realistic simulation, only the heading component of the kinematic observation is utilized in our simulation, as it represents information that would be readily available from an onboard IMU sensor.

\paragraph{2) LiDAR Observation.}
The LiDAR observation models the ego vehicle’s $360^\circ$ surroundings using a fixed number of angular beams, each returning the distance and relative velocity of the nearest obstacle. The LiDAR observation is represented as an $N \times 2$ array:
\begin{equation}
o^{\text{LiDAR}}_t = [d_j,\, \dot{d}_j]_{j=1}^{N},
\end{equation}
where $d_j$ and $\dot{d}_j$ denote the range and radial velocity along the $j^{\text{th}}$ beam, respectively. This modality provides a spatially and dynamically rich perception of the surrounding environment, facilitating awareness of motion and proximity in all directions.

\paragraph{3) Visual Observation (BEV Mode).}
In the bird’s-eye-view (BEV) mode, the environment generates a top-down grayscale or RGB image centered on the ego vehicle. This visual input captures lane markings, road geometry, and the positions of nearby vehicles, providing high-level spatial context ideal for convolutional or transformer-based perception models.

\subsection{Action Space}
The agent interacts with the environment through a discrete \textit{meta-action space} that abstracts high-level driving intentions rather than low-level continuous control commands. The available action set is defined as:
\begin{equation}
\mathcal{A} = \{\text{LANE\_LEFT},\; \text{IDLE},\; \text{LANE\_RIGHT},\; \text{FASTER},\; \text{SLOWER}\}.
\end{equation}
Each action represents a tactical driving maneuver, such as changing lanes or adjusting longitudinal speed. Invalid actions, such as lane changes beyond the road boundary, are automatically replaced with \textit{IDLE}. This discrete formulation simplifies the decision-making process by focusing on strategic behaviors such as overtaking, merging, and maintaining safe following distances, without modeling detailed vehicle dynamics directly.

\subsection{Highway Scenario}
The \textit{Highway scenario} in \texttt{Highway-Env} serves as a standard benchmark for high-level decision-making in autonomous driving. It models a multi-lane straight highway with surrounding traffic, where the ego vehicle is required to maintain high speed, avoid collisions, and perform safe lane changes. The default configuration includes four lanes, approximately fifty vehicles, a simulation frequency of $15~\text{Hz}$, and a policy frequency of $1~\text{Hz}$. Each episode lasts for $40~\text{s}$ or terminates earlier upon collision.

The default reward function encourages efficient and safe driving while penalizing collisions. A common formulation is expressed as
\begin{equation}
r_t = \alpha \frac{v_t - v_{\min}}{v_{\max} - v_{\min}} - \beta\, \mathbb{I}_{\text{collision}},
\end{equation}
where $v_t$ denotes the ego vehicle’s longitudinal velocity, $\mathbb{I}_{\text{collision}}$ is an indicator function that equals one upon collision, and $\alpha, \beta > 0$ are weighting coefficients. The default parameters are $v_{\min} = 20~\text{m/s}$, $v_{\max} = 30~\text{m/s}$, and $\alpha = \beta = 1$. 

\subsection{Roundabout Scenario}\label{app:A:round}
The \textit{Roundabout scenario} in \texttt{Highway-Env} is designed to evaluate autonomous driving policies in complex interactive traffic situations involving merging, yielding, and priority management. The environment models a multi-lane circular intersection (roundabout) with multiple entry and exit points. The ego vehicle must safely merge into the circulating traffic, navigate the roundabout, and exit through the designated lane while avoiding collisions and unnecessary delays. This setup provides a more challenging scenario compared to the straight highway, as it requires coordination with other vehicles and understanding of right-of-way dynamics.

The default configuration includes several controlled and uncontrolled entry branches, dynamic traffic flow, a simulation frequency of $15~\text{Hz}$, and a policy frequency of $1~\text{Hz}$. Each episode terminates upon collision, successful exit, or after $40~\text{s}$ of simulation time.

The default reward function is designed to promote smooth, safe, and goal-oriented navigation. It incorporates terms for velocity maintenance, collision avoidance, and lane-change efficiency. A representative formulation is:
\begin{equation}
r_t = \alpha \frac{v_t - v_{\min}}{v_{\max} - v_{\min}} 
      - \beta\, \mathbb{I}_{\text{collision}}
      + \gamma\, \mathbb{I}_{\text{lane-change}}
      + \delta\, \mathbb{I}_{\text{on-road}},
\end{equation}
where $v_t$ is the ego vehicle’s longitudinal velocity, $\mathbb{I}_{\text{collision}}$ and $\mathbb{I}_{\text{on-road}}$ are binary indicators for collision and road keeping events, respectively, and $\mathbb{I}_{\text{lane-change}}$ represents a successful or efficient lane-change maneuver. The coefficients $\alpha$, $\beta$, $\gamma$, and $\delta$ balance the importance of speed, safety, and maneuver efficiency. Default parameter values are $\alpha=0.2$, $\beta=1.0$, $\gamma=0.05$, and $\delta=0.5$. These parameters ensure that after each action, the maximum achievable reward is $1$.  

\section{LiDar to Image Conversion}\label{app:B}
To convert raw LiDAR measurements into a structured spatial representation suitable for learning-based decision making, we construct an image-like representation of LiDAR data that encodes both spatial and kinematic information surrounding the ego vehicle. 

Each LiDAR scan provides a set of distance--velocity pairs 
\(\{(r_i, v_i)\}_{i=1}^{N}\) measured along discrete azimuth angles \(\phi_i\). 
For each valid beam, we uniformly sample intermediate points along the ray as
\begin{equation}
d_{i,k} \in [r_i, 1] \cdot d_{\max}, 
\qquad k = 0, \ldots, K_i,
\end{equation}
and project them onto the ground plane:
\begin{equation}
x_{i,k} = d_{i,k} \cos\phi_i, 
\qquad 
y_{i,k} = d_{i,k} \sin\phi_i .
\end{equation}

The LiDAR-sensed \textit{radial velocity} \(v_i\) is combined with the ego vehicle's forward velocity \(v_{\text{ego}}\)
to compute a \textit{decayed velocity intensity}:
\begin{equation}
z_{i,k} = (0.98)^k \, (v_i + v_{\text{ego}}),
\end{equation}
which decreases with increasing distance due to the unavailable data beyond $r_i$.
Each point \((x_{i,k}, y_{i,k}, z_{i,k})\) is discretized according to the voxel size 
\(\Delta x = \Delta y = \text{voxel\_size}\), forming a velocity-weighted BEV map \(\mathcal{G}(x,y)\).

To explicitly represent the ego vehicle, we overlay a rotated rectangular footprint centered in the grid. 
Given the vehicle heading angle \(\theta\), grid coordinates are rotated as:
\begin{equation}
\begin{bmatrix}
x'\\[3pt]
y'
\end{bmatrix}
=
\begin{bmatrix}
\cos\theta & \sin\theta\\[3pt]
-\sin\theta & \cos\theta
\end{bmatrix}
\begin{bmatrix}
x - c_x\\[3pt]
y - c_y
\end{bmatrix},
\end{equation}
and all cells satisfying 
\(|x'| \leq L/2\) and \(|y'| \leq W/2\) are assigned the ego velocity value \(v_{\text{ego}}\), and $L$ and $W$ are ego vehicle's dimensions.

Finally, the provided grid map is normalized, 
and mapped to a color scale to generate an RGB BEV image:
\begin{equation}
\text{RGB}(x,y) = \text{colormap}\!\left( \frac{\mathcal{G}(x,y)}{s} \right),
\end{equation}
where \(s\) is a scaling constant to normalize the velocity values ($s=v_{max}$).
The resulting tensor \(\text{RGB} \in \mathbb{R}^{3 \times H \times W}\) 
provides a compact visual encoding of the \textbf{spatial occupancy} and 
\textbf{local velocity field} surrounding the ego vehicle, 
which is used as the sensory input for downstream reinforcement learning and perception modules.

\section{Experimental Settings}
\label{app:c}

\subsection{Sensor and Reward Function Configurations}

In all simulation scenarios of the \textit{Highway-Env} environment, different perceptual fields of view were defined for the visual and LiDAR modalities to reflect their complementary sensing capabilities. 
The grayscale BEV representation was constructed with a field of view of approximately $35\,\text{m}$ around the ego vehicle, providing detailed short- to mid-range spatial awareness and lane-level interaction information. 
In contrast, the LiDAR modality was configured with an extended sensing range of $60\,\text{m}$, enabling the perception of long-range structural and dynamic information from surrounding traffic participants. 
This asymmetric configuration allows the BEV input to emphasize fine-grained spatial features close to the ego vehicle, while the LiDAR input captures broader environmental context and motion dynamics essential for anticipatory decision-making. 

It should be noted that extending the BEV field of view beyond approximately $40$–$50\,\text{m}$ is often impractical or unreliable in both simulated and real-world settings due to the inherent limitations of image-based BEV generation, such as reduced pixel density, camera perspective distortion, and increased occlusion at longer distances~\cite{zhao2024bev,bloesbev2024}. 
Consequently, the choice of a $35\,\text{m}$ BEV range aligns well with realistic perception constraints observed in camera-based BEV systems, which typically operate effectively within a range of $30$–$50\,\text{m}$~\cite{zhao2024bev}. 
For the single-modal versus multi-modal comparison experiments, the same $35\,\text{m}$ field of view was applied to both the LiDAR and image inputs to ensure a fair and controlled comparison across sensing modalities. 
All experiments were conducted using the default parameter settings of the environment’s reward function. 

\subsection{Model Architecture and Training Hyperparameters}
\label{app:hyperparameter}
The multi-modal architecture and its hyperparameters are given in Table \ref{tableapp}.  

\begin{table*}[h!]
\centering
\caption{Multi-Modal Transformer architecture and reinforcement learning configuration.}
\label{tab:architecture_hyper}
\begin{tabular}{@{}lcc@{}}
\toprule
\textbf{Parameter} & \textbf{Symbol / Setting} & \textbf{Value / Description} \\ 
\midrule
\multicolumn{3}{c}{\textbf{Feature Extraction (per modality)}} \\ 
\midrule
Number of convolutional layers & -- & 3 × Conv2D per modality \\
\textbf{BEV kernel sizes} & -- & [5×5, 3×3, 3×3] \\
\textbf{LiDAR kernel sizes} & -- & [7×7, 5×5, 3×3] \\
Stride & -- & [3, 2, 1] / [3, 3, 1] \\
Channels per layer & -- & 8 → 16 → 16 \\
Activation function & -- & ReLU / Binary LIF \\
Output embedding dimension & $d_{\text{model}}$ & 32 \\
\midrule
\multicolumn{3}{c}{\textbf{Cross-Attention Fusion Module}} \\ 
\midrule
Number of attention heads & $N_h$ & 8 \\
Feed-forward dimension & $d_{\text{ff}}$ & 128 \\
Positional encoding & -- & Learnable Positional Encoding \\
Normalization & -- & LayerNorm \\
\midrule
\multicolumn{3}{c}{\textbf{Decision Head}} \\ 
\midrule
Activation function & -- & ReLU / Binary LIF \\
Feed-forward dimension & $d_{\text{ff}}$ & 512 \\
Output dimension & -- & 5 (discrete actions in \textit{Highway-Env}) \\
\midrule
\multicolumn{3}{c}{\textbf{Reinforcement Learning Parameters}} \\ 
\midrule
Algorithm & -- & Deep Q-Network (DQN) \\
Discount factor & $\gamma$ & 0.99 \\
Replay buffer size & -- & $5\times10^{4}$ transitions \\
Batch size & -- & 64 \\
Target network update frequency & -- & every 100 steps \\
Learning rate & $\eta_0$ & $1\times10^{-4}$ \\
Optimizer & -- & Adam \\
Exploration schedule & $\epsilon$-greedy & $\epsilon$ decays linearly from 1.0 to 0.1 over $7\times10^{4}$ steps \\
Reward weights & -- & Default \textit{Highway-Env} (speed, collision, lane-change) \\
Training steps per scenario & -- &  $1\times10^{5}$\\
Evaluation episodes & -- & 20-50 \\
\multicolumn{3}{c}{\textbf{Spiking-Specific Parameters}} \\ 
\midrule
Membrane time constant ($\tau_m$) & -- & 2 \\
Spike threshold (positive) ($V_{\text{th}}^{+}$) & -- & 1.0 \\
Spike threshold (negative) ($V_{\text{th}}^{-}$) & -- & $-4$ \\
Reset mechanism ($V_{\text{reset}}$) & Subtractive reset ($V \leftarrow V - V_{\text{th}}$) & Subtractive reset ($V \leftarrow V - V_{\text{th}}^{\pm}$) \\
Output spikes &--& Binary $\{0, +1\}$, Ternary $\{-1, 0, +1\}$ \\
Surrogate gradient type & -- & Arctangent \\
Simulation time window ($T_s$) &--& 5 \\
Neuron model &--& binary LIF and ternary LIF (asymmetric thresholds) \\
\bottomrule
\end{tabular}
\label{tableapp}
\end{table*}

The spiking architectures were implemented using the \textbf{SNNtorch} library~\cite{eshraghian2021training}, which provides a PyTorch-based framework for constructing and training temporal spiking neural networks. 
For the ternary neuron design, we adopted the asymmetric ternary LIF proposed in ~\cite{ghoreishee2025improving}, where the neuron employs two different firing thresholds—one positive $(V_{\text{th}}^{+})$ and one negative $(V_{\text{th}}^{-})$—to generate ternary spike outputs $\{-1, 0, +1\}$. 
In our implementation, we initially used symmetric thresholds of equal magnitude, but empirical evaluations confirmed that the \textit{asymmetric configuration}, as suggested in~\cite{ghoreishee2025improving}, yielded improved convergence and higher reward consistency across driving scenarios.

\subsection{Energy Consumption Analysis}

SNNs have attracted substantial attention due to their inherently low power consumption and biologically inspired computing paradigm. This efficiency stems from their \textit{event-driven nature}, where neurons and synapses are activated only upon spike events, drastically reducing redundant computations. In a spiking-driven matrix, where elements are binary (0 or 1), element-wise multiplication degenerates into a simple masking operation, resulting in negligible energy cost. Moreover, matrix multiplications in SNNs can be reformulated as sparse accumulation operations, which are efficiently implemented on neuromorphic hardware through \textit{address-event representation (AER)} or other addressable addition techniques~\cite{frenkel2023bottom}.

The total energy consumption of artificial and spiking networks can be estimated as
\begin{equation}
E_{\text{ANN}} = \text{FLOPs} \times E_{\text{MAC}},
\label{eq:EANN}
\end{equation}
\begin{equation}
E_{\text{SNN}} = \text{FLOPs} \times E_{\text{AC}},
\label{eq:ESNN}
\end{equation}
where $E_{\text{MAC}}$ denotes the energy required for a multiply-accumulate (MAC) operation in a conventional artificial neural network (ANN), and $E_{\text{AC}}$ represents the energy cost of an accumulate (AC) operation in an SNN. Following the empirical measurements reported in~\cite{horowitz20141}, we can use $E_{\text{MAC}} = 4.6~\text{pJ}$ and $E_{\text{AC}} = 0.9~\text{pJ}$.

To ensure a comprehensive estimation of computational cost, the FLOPs calculation should incorporate both the \textit{spike firing rate} and the \textit{total simulation time steps} of the spiking neurons. This approach enables a more accurate comparison of the energy efficiency between SNN-based and ANN-based architectures. In our design, the total simulation time window was set to 5 to ensure both energy efficiency and acceptable latency for decision-making. 

By assuming that the number of operations remains nearly identical between the spiking and non-spiking models and considering only 5 simulation time steps, we can estimate the energy efficiency margin based on the observed spike activity. As shown in Fig.~\ref{Result7}, the worst-case spike density in the SNN variants is approximately 17\%. According to the theoretical model, this implies an energy cost of about 
$E_{\text{SNN}} \approx 5 \times 0.17 \times 0.9 = 0.77~\text{pJ}$ 
per operation, compared to 
$E_{\text{ANN}} = 4.6~\text{pJ}$. 
This demonstrates a substantial energy efficiency advantage for the spiking variant. 

Furthermore, as illustrated in Fig.~\ref{Result7}, the proposed TTSA-based model exhibits approximately 40\% higher spike sparsity than the SSA-based model. Since the number of operations remains identical between the two architectures, the theoretical energy model suggests that the TTSA-based SNN reduces energy consumption by roughly 40\% relative to the SSA-based counterpart.
\end{document}